# Object-Oriented Bayesian Networks

**Daphne Koller**
Stanford University
koller@cs.stanford.edu

**Avi Pfeffer**
Stanford University
avi@cs.stanford.edu

## Abstract

Bayesian networks provide a modeling language and associated inference algorithm for stochastic domains. They have been successfully applied in a variety of medium-scale applications. However, when faced with a large complex domain, the task of modeling using Bayesian networks begins to resemble the task of programming using logical circuits. In this paper, we describe an object-oriented Bayesian network (OOBN) language, which allows complex domains to be described in terms of inter-related objects. We use a Bayesian network fragment to describe the probabilistic relations between the attributes of an object. These attributes can themselves be objects, providing a natural framework for encoding part-of hierarchies. Classes are used to provide a reusable probabilistic model which can be applied to multiple similar objects. Classes also support inheritance of model fragments from a class to a subclass, allowing the common aspects of related classes to be defined only once. Our language has clear declarative semantics: an OOBN can be interpreted as a stochastic functional program, so that it uniquely specifies a probabilistic model. We provide an inference algorithm for OOBNs, and show that much of the structural information encoded by an OOBN—particularly the encapsulation of variables within an object and the reuse of model fragments in different contexts—can also be used to speed up the inference process.

## 1 Introduction

Over the past decade, Bayesian networks [Pearl, 1988] have established themselves as an effective and principled framework for knowledge representation and reasoning under uncertainty. The sound probabilistic semantics, explicit encoding of relevance relationships, and inference algorithms that are fairly efficient in practice, have led to the use of Bayesian networks in a wide variety of applications.

Despite their great success, Bayesian networks (BNs) are inadequate as a general knowledge representation language for large and complex domains [Mahoney and Laskey, 1996]. In a traditional BN, each node corresponds to some basic attribute of the domain. The set of nodes and the network structure are fixed in advance, so that the network can only be used in the specific domain for which it was created. The construction of a network is a painstaking manual process, somewhat analogous to programming using logical circuits. For example, the only mechanism for "code reuse" is manually copying a network fragment and pasting it somewhere else. This manual duplication has the same drawbacks as the analogous process in a programming task. For example, if the original network fragment is modified, the knowledge engineer must manually go back and change all of the models which used it.

These difficulties have been encountered and largely solved in the context of programming languages, primarily via the introduction of abstract data types. *Object-oriented* programming languages [Goldberg and Robson, 1983] provide a framework for organizing abstract data types in a way that allows for robust, flexible and efficient construction of programs. Similarly, object-oriented database systems [Banerjee *et al.*, 1987] provide tools for managing rich, complex data. In this paper, we present *object-oriented Bayesian networks (OOBNs)*, a powerful and general framework for large-scale knowledge representation using Bayesian networks. As we will show, OOBNs combine clear declarative probabilistic semantics with many of the organizational benefits of an object-oriented framework. We will also show that object-oriented models help reveal the locality structure of a domain, thereby supporting more efficient probabilistic inference.

The basic element in an OOBN is an object. The most basic object is a standard random variable, as in traditional BNs. However, an OOBN also has more complex objects. In general, an object has some set of attributes, each of which is an object. Thus, a car object may have several attributes, such as the car's color, its owner, its engine, etc. The car's color is a simple object, taking values in some finite range, while the car's owner is itself a complex object with its own attributes.

The value of an object is an assignment of values to all of its attributes. We use Bayesian networks to define a probabilistic model over the assignments of values to an object. As usual, this probabilistic model must take into account the influences of the environment on the object. Based on our work in [Koller *et al.*, 1997b], we view each object as a *stochastic function* from its inputs—the attributes which influence it—to its outputs. Briefly, a stochastic function is a function that, for each value of its inputs, returns a probability distribution over the value of its outputs. A conditional probability table (CPT) in a traditional Bayesian network is a simple kind of stochastic function, that can be used to define a simple object—a basic random variable in a BN. Stochastic functions can be composed, and more complex functions can be defined in terms of simpler ones, as in a functional programming language. A complex object is defined by assigning stochastic functions to each of its attributes, and connecting the attributes in a Bayesian network. The result is a stochastic function for the complex object. A very simple version of this representation was used by Srinivas [Srinivas, 1994] in the context of model-based fault diagnosis in a hierarchical component model.

OOBNs allow us to generalize over multiple objects. Formally, we define *classes* of objects, all of which are described using the same probabilistic model. Classes provide the ability to describe a general, reusable model that can be used in many different contexts. The encapsulation of the internal details of a class description by the class interface allows classes to be used as libraries and combined into a model as needed. It is important to realize that OOBNs are *not* simply a library of C++ classes each of which constructs a Bayesian network fragment. Models defined in OOBNs have clear declarative probabilistic semantics: just like a BN, an OOBN uniquely specifies a probability distribution. Thus, an OOBN can be interpreted unambiguously, without having to understand some associated infrastructure of procedural model construction code.

In addition to supporting generalization, classes serve another important role. In general, many different classes share common substructures. For example, while we may have a general class for people, we may also want a more specific class for college students, with a probabilistic model describing their performance as students based on their background and abilities. Clearly, the two models are not completely disjoint, since students also have all of the attributes used to describe a person, probably described using the same probabilistic model. The class mechanism in OOBNs allows the student class to inherit much of its structure from the person class, modifying or augmenting it where necessary. The stochastic function perspective helps provide a clean semantics: since each attribute in a class is a stochastic function, a subclass simply redefines some of the functions and/or adds some new ones.

We can view the inheritance hierarchy over classes as supporting an is-a hierarchy over objects. The ability to enclose objects within other objects (as values of attributes) provides an orthogonal part-of hierarchy. The combination of these two hierarchies provides a natural framework for dynamic abstraction and refinement of models. In many cases, we can view a class as a more abstract version of its subclasses, one which ignores some (less important) details. For example, we can *iconize* a class, compiling away all of the details about its internal structure. As inference proceeds, the user can decide to refine the model by using a more specific class for one or more of the objects in the model. At the same time, the user can focus in along the part-of hierarchy to refine those parts of a model that are most relevant.

OOBNs are more than just a nice language for representing complex probabilistic models. By representing the domain as a hierarchy of interconnected objects, an OOBN makes explicit certain organizational structure. As we have seen many times before, by making additional structural information accessible to the inference algorithm, we can significantly improve the performance of inference.

Specifically, in the network defined by an OOBN model, the internal parts of an object are encapsulated within the object. Probabilistically, this implies that the encapsulated attributes are d-separated from the rest of the network by the object's inputs and outputs. This separation property can be utilized to localize probabilistic computation within objects, with only a limited interaction between them. The *multiply sectioned Bayesian network (MSBN)* framework of Xiang *et al.* [Xiang *et al.*, 1993] turns out to be a particularly appropriate mechanism for utilizing this structure.

Since objects of the same class have the same probabilistic model, we can precompute certain parts of the inference task on the level of the class, and reuse them for different instances. Localization and reuse of computation also provide advantages during the refinement process. When a particular part of a model is refined, the effects of that refinement can be computed locally within the refined object, and propagated (but only if necessary) through the object's interface. Furthermore, if the new class for the object inherits parts of its definition from its superclass, our algorithm can simply reuse its previous computation for those parts that stay the same.

## 2 The object-oriented framework

### 2.1 Objects and types

The basic unit of discourse in an OOBN is the *object*. One way of viewing an object is as a collection of properties that are associated with some entity in our domain. An object will sometimes correspond to a physical entity in the world being modeled, but it may also represent an abstract entity, or a relationship between different entities. For example, a vehicle surveillance model may use objects such as cars, drivers and roads, that correspond to physical entities. The objects in a medical diagnosis model may correspond to more abstract entities such as diseases and symptoms, as well as physiological systems such as the respiratory system. We begin with some basic type definitions, which will form the foundation for the type system of our language.

**Definition 2.1:** A *basic type* is a set of values, which is either one of the predefined basic types—Booleans, Integers or Reals—or some user-defined enumerated set (e.g., WEATHER = {*raining, cloudy, sunny*}).

A *basic variable* is a variable which takes values in some basic type. ∎

A structured type is defined recursively out of the basic types.

**Definition 2.2:** A *structured type* is a set of values defined by a tuple $\langle A_1 : t_1, \ldots, A_n : t_n \rangle$, where $A_1, \ldots, A_n$ are *attribute labels* and $t_1, \ldots, t_n$ are corresponding (basic or structured) types. The set of values of this type are all those of the form $\langle A_1 : v_1, \ldots, A_n : v_n \rangle$ where each $v_i$ is of type $t_i$. A *structured variable* is a variable which takes values in some structured type. ∎

For any variable $V$, let $Val(V)$ denote the set of values that $V$ can take. ($Val(\mathbf{V})$ for a set of variables $\mathbf{V}$ is defined in the obvious way.) If $v = \langle A_1 : v_1, \ldots, A_k : v_k \rangle$ is a value for $V$, then $v.A_i$ refers to $v_i$.

**Example 2.3:** A structured type representing a person may have the attributes *Age*, *Gender*, *Description*, and *Income*. The *Age* attribute has a basic user-defined type AGES = {*0-20yr, 21–30yr, 31–60yr, 60+yr*}. The *Gender* attribute also has a user-defined type GENDERS = {*male, female*}. *Income* takes values in the user defined type INCOMES = { *$0–10K, $10–30K, $30–80K, $80K+* }. The *Description* attribute has a structured type, with attributes *Hair-Color*, *Eye-Color*, and *Height*, each of which takes values in some appropriately defined type. One possible value $v$ of this type may be $\langle$Age : 21–30yr, Gender : female, Income: $80K+, Description : $\langle$Hair-Color : brown, Eye-Color : green, Height : 5ft8in$\rangle\rangle$. The value of *v.Gender* is *female* and the value of *v.Description.Height* is *5ft8in*. ∎

Given these basic definitions, we can now define objects. Objects are composed of two types of attributes: *input attributes* and *value attributes*. Input attributes are parameters to the object; their value influences the choice of values for the encapsulated and output attributes. Value attributes are part of the specification of the object itself. Complex objects are created when these attributes take other objects as values. Value attributes are divided into two types: *output attributes*, and *encapsulated attributes*. The former are visible to the rest of the model, whereas the latter are encapsulated within the object.

**Definition 2.4:** A *simple object* $X$ is composed of a set of labelled *input attributes* $I_1, \ldots, I_k$ and a single *output attribute* labelled *Output*. All of $X$'s attributes are basic variables. ∎

Intuitively, a simple object corresponds to a random variable in a Bayesian network. It takes on values in some simple type, and depends on other random variables (its inputs). Complex objects are composed of simple objects.

**Definition 2.5:** A *complex object* $X$ is composed of a set of labelled *attributes*. The attributes are partitioned into three sets: the *input attributes* $\mathcal{I}(X)$, the *output attributes* $\mathcal{O}(X)$, and the *encapsulated attributes* $\mathcal{E}(X)$. The output attributes and encapsulated attributes are called *value attributes*, and denoted $\mathcal{A}(X)$. The input attributes are (basic or structured) variables. The value attributes are themselves objects. ∎

Note that input attributes are not objects. As we will see later, they correspond to "parameters" passed by value to the enclosing object $X$.

We use $X.A_i$ to denote the object or variable represented by the attribute $A_i$ of an object $X$. More generally, we can define the object or variable $X.\rho$ for an attribute chain $\rho = A_1.A_2.\cdots.A_k$.

An object can be viewed as defining two structured variables, one corresponding to its full value, and one only to the part which is visible to the rest of the objects in the model. For simple objects, these two variables are the same.

**Definition 2.6:** For a simple object $X$, its *full variable* $\vartheta^+(X)$ and its *output variable* $\vartheta(X)$ are defined to be $X$'s output variable $X.Output$. ∎

**Definition 2.7:** For a complex object $X$, its *full variable* $\vartheta^+(X)$ is a structured variable whose value is $\langle A_1 : V_1, \ldots, A_n : V_n \rangle$, where $A_1, \ldots, A_n$ are $X$'s value attributes, and $V_i$ is $\vartheta^+(X.A_i)$. Analogously, the *output variable* $\vartheta(X)$ is a structured variable whose value is $\langle O_1 : U_1, \ldots, O_k : U_k \rangle$, where $O_1, \ldots, O_k$ are $X$'s output attributes and $U_i$ is $\vartheta(X.O_i)$. ∎

Clearly, $\vartheta(X)$ is a projection of $\vartheta^+(X)$.

The distinction between objects and the variables they define is important to ensure clean semantics of our language. Objects are organizational units, describing the entities in the domain and the relationships between them. The variables defined by objects correspond to actual random variables in a probabilistic model. The variable corresponding to an object defines its type.

**Definition 2.8:** A *value type* of an object $X$ is the type of $\vartheta^+(X)$. The *output type* of $X$ is the type of $\vartheta(X)$. ∎

**Example 2.9:** A complex object $X$ representing a car may be composed of the following attributes: $\mathcal{I}(X)$ contains a single attribute *Owner* which is a variable of structured type PERSON, as described in Example 2.3. $\mathcal{E}(X)$ consists of the attributes *Original-Value*, *Age*, *Mileage*, *Type*, and *Maintenance*, each of which is a simple object, and the attributes *Tires*, *Engine*, *Brakes* and *Steering*, each of which is a complex object. The (output or value) type of *Original-Value* is the user-defined type {*$0-10K, $10–20K, $20K+*}. The output type of *Engine* is $\langle$*Reliability* : {*good, bad*}, *Power* : {*high, medium, low*}$\rangle$. $\mathcal{O}(X)$ consists of the attributes *Current-Value*, *Max-Speed*, *Braking-Power* and *Steering-Safety*, all of which correspond to simple objects. One value of $\vartheta(X)$ may be $\langle$*Current-Value* : *$0-10K*, *Max-Speed* : *70mph*, *Braking-Power* : *medium*, *Steering-Safety* : *good*$\rangle$. A value for $\vartheta^+(X)$ would, of course, be much more complex. ∎

## 2.2 Stochastic functions

Since we are interested in modeling our uncertainty about the possible values of an object, we associate each object with a probabilistic model. The model defines a distribution over the object's value as a function of the values of its input attributes. More precisely, we associate a *stochastic function* [Koller *et al.*, 1997b] with the object that defines, for each assignment of values to the object's inputs, a distribution over the possible values for the object.

**Definition 2.10:** Let $t = t_1, \ldots, t_k$ and $u$ be value types. A *stochastic function* from $t$ to $u$ is a function from $Val(t)$ to probability distributions over $Val(u)$. ∎

In our representation, we describe stochastic functions using recursive composition of *object-oriented (Bayesian) network fragments (OONF)*, each of which specifies a conditional distribution of a set of value attributes given some set of input attributes.

**Definition 2.11:** A *simple object-oriented network fragment* $F$ has a set of input attributes $\mathcal{I}(F)$ and a single value attribute *Output*, all of which are basic variables. The network consists of a conditional probability function defining a distribution over $Val(Output)$ for each assignment of values in $Val(\mathcal{I})$. ∎

A simple OONF can be represented as a conditional probability table (CPT), as in most Bayesian networks. Often, however, a more compact representation is appropriate (particularly when one of the attributes takes values in an infinite space such as the integers or reals). In the ensuing discussion, we assume simple OONFs are represented by CPTs, but the same arguments hold for any representation.

**Definition 2.12** An *object-oriented network fragment* $F$ over the input attributes $\mathcal{I}(F)$ and the value attributes $\mathcal{A}(F)$ consists of a directed acyclic graph whose nodes are the attributes of $F$, and, for each value attribute $A \in \mathcal{A}(F)$:

- For each input $I$, an annotation $B.\rho$, where $B$ is a parent of $A$ and $\rho$ is an attribute of $\vartheta(B)$. We require that the attributes $A.I$ and $\vartheta(B).\rho$ have the same type. We also require that every parent of $A$ be used to annotate at least one input of $A$.

- An OONF $F_A$. If $A$ is a simple attribute, then $F_A$ must be a simple OONF. ∎

Intuitively, the annotations on the inputs of an attribute $A$ represent the exact mapping between the parameters to $A$ and the values which are passed to them.

Figure 1(a) demonstrates the DAG for the probabilistic model associated with a CAR object. We use the exact endpoints of the edges to represent the annotation on the edges. Thus, for example, one of the input attributes of the *Max-Speed* attribute is mapped to the output variable of the simple object *Type*, while the other is mapped to the attribute $\vartheta(Engine).Power$. The input attributes are not shown since they are usually clear from context.

An OONF $F$ encodes a certain set of assumptions about the probabilistic model over variables defined by the objects in the OONF:

1. **Conditional independence assumptions:** For each value attribute $A$ in $F$, and each attribute $B$ in $F$ which is a non-descendant of $A$, $\vartheta^+(A)$ is conditionally independent of $\vartheta^+(B)$ given a value for $\mathcal{I}(A)$.

2. **Equality assumptions:** For each input $I$ of attribute $A$ annotated with $B.\rho$, $A.I = \vartheta(B.\rho)$.

3. **Distribution assumptions:** The conditional distribution over $\vartheta^+(A)$ given $\mathcal{I}(A)$ is defined via $F_A$.

**Theorem 2.13:** *Under the assumptions (1)–(3), an OONF $F$ over $\mathcal{I}(F)$ and $\mathcal{A}(F)$ uniquely specifies a stochastic function from (the type of) $\mathcal{I}(F)$ to (the type of) $\mathcal{A}(F)$.*

**Proof:** The proof proceeds by induction on the depth of the OONF, where the depth of a simple OONF is 0, and the depth of a complex OONF is defined to be the maximal depth of the OONFs for its attributes plus 1. The base case is trivial: a simple OONF is a CPT, which clearly defines a conditional distribution of the type that we want.

The inductive step essentially uses a very similar chain rule to the one used for standard Bayesian networks [Pearl, 1988]. Let $A_1, \ldots, A_n$ be the value attributes in $\mathcal{A}(F)$, ordered in a way which is compatible with the DAG defined by $F$. Fix a value for the variables in $\mathcal{I}(F)$, and consider some assignment of values $v_1, \ldots, v_n$ to $\vartheta^+(A_1), \ldots, \vartheta^+(A_n)$. By simple probabilistic reasoning, we have that

$$P(\vartheta^+(A_1) = v_1, \ldots, \vartheta^+(A_n) = v_n) = \\ \prod_i P(\vartheta^+(A_i) = v_i \mid \vartheta^+(A_1) = v_1, \ldots, \vartheta^+(A_{i-1}) = v_{i-1}).$$

Now, consider a single term in the product $P(\vartheta^+(A_i) = v_i \mid \vartheta^+(A_1) = v_1, \ldots, \vartheta^+(A_{i-1}) = v_{i-1})$, and let $\boldsymbol{A}$ denote $A_1, \ldots, A_{i-1}$ and $\boldsymbol{v}$ denote $v_1, \ldots, v_{i-1}$. Since the DAG is acyclic and the ordering on variables is compatible with the DAG, all of the full variables corresponding to $A_i$'s parents are assigned values in the right hand side of the conditional probability. The equality assumption (2) therefore implies that all of $A_i$'s input attributes are also assigned unique values. Thus, using the conditional independence assumption (1), there is some unique value $\boldsymbol{u}$ such that

$$P(\vartheta^+(A_i) = v_i \mid \vartheta^+(\boldsymbol{A}) = \boldsymbol{v}) = \\ P(\vartheta^+(A_i) = v_i \mid \vartheta^+(\boldsymbol{A}) = \boldsymbol{v}, \mathcal{I}(A_I) = \boldsymbol{u}) = \\ P(\vartheta^+(A_i) = v_i \mid \mathcal{I}(A_i) = \boldsymbol{u}).$$

Finally, the distribution assumption (3) implies that the conditional distribution for $A_i$—$P(\vartheta^+(A_i) \mid \mathcal{I}(A_i))$—is defined via the OONF $F_A$. The inductive hypothesis applied to $F_A$ implies the existence of a unique stochastic function for $\vartheta^+(A_i)$ given $\mathcal{I}(A_i)$.

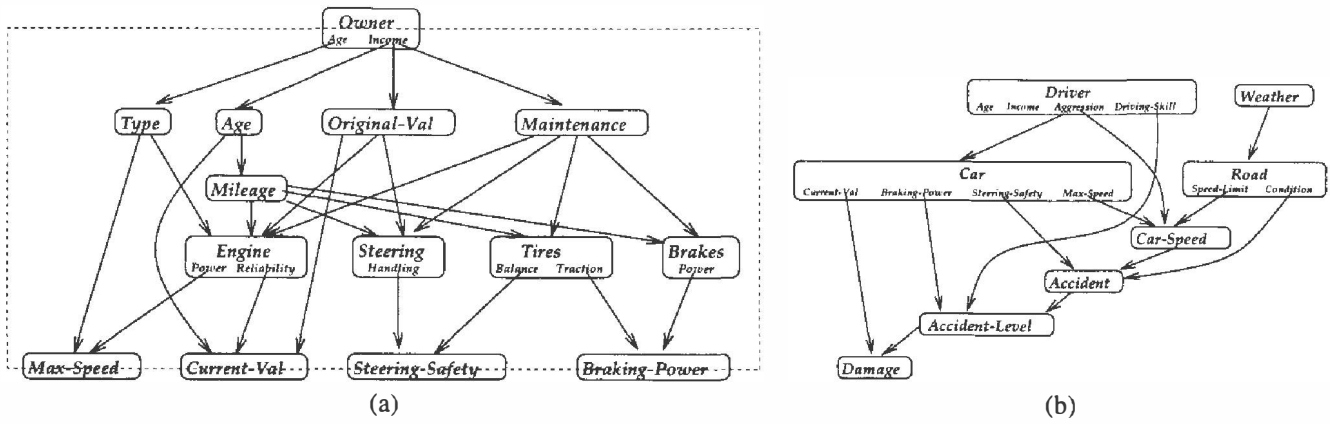

Figure 1: (a) A graphical depiction of the interconnection model for $M_{car}$. (b) OOBN model for a car accident.

Thus, the value of each term in the product is uniquely defined, so that the entire conditional distribution is also uniquely defined, as required. ∎

### 2.3 Classes and OOBNs

A stochastic function, as described in the previous section, describes the probabilistic relation between a set of input attributes and a set of value attributes. We can associate a stochastic function with an object whose type matches the type of the stochastic function. More precisely, an object $X$ is *type-compatible* with an OONF $F$ if $F$ is an OONF over the input attributes $\mathcal{I}(X)$ and the value attributes $\mathcal{A}(X)$.

Recall that the OONF for $X$ would also have to specify an OONF for each of $X$'s attributes. In general, complex models often involve many similar objects (or attributes of objects), whose stochastic functions are essentially identical. Therefore, we would like to be define generic OONFs, which can be used multiple times in defining many similar objects. We accomplish this goal using the notion of a *class*. A class $C$ is simply an OONF which is not associated with any specific object. We can associate an OONF with an object simply by asserting that the object is of class $C$. Clearly, we can also associate the same OONF with multiple objects.

We are now ready to define an object-oriented Bayesian network. An OOBN is essentially a single *situation* object with no inputs, whose probabilistic properties are defined using an associated OONF.

**Definition 2.14:** An *object-oriented Bayesian network* consists of a set of class definitions $C_1, \ldots, C_m$, and a single *Situation* object, which has no input attributes, with an associated OONF. An object $X$ in the model (including within a class definition) can only be associated with one of the defined classes $C_i$ ($i = 1, \ldots, m$), and $C_i$ must be type-compatible with $X$. ∎

Note that the type-compatibility requirement, which incorporates all of the value attributes of an object, ensures that class definitions are non-recursive.[1] The value type of an attribute $X.A$ must be strictly simpler than the value type of $X$. Therefore, if a class $C'$ is used to specify an OONF for an attribute $A$ within a class $C$, then $C$ cannot be used (even indirectly) within $C'$. Thus, we can translate an OONF incorporating class definitions into an OONF without such definitions simply by "unrolling" class definitions. Combining this observation with Theorem 2.13, we obtain the following result:

**Theorem 2.15:** *An OOBN $\mathcal{B}$ uniquely specifies a probability distribution over $\vartheta^+(Situation)$.*

**Example 2.16:** Let us outline the construction of a simple OOBN model for a car accident. Our OOBN will contain class definitions for the following classes: CAR, ENGINE, STEERING, TIRES, BRAKES, WEATHER, ROAD, and DRIVER. The internal structure of the OONFs for the CAR class is shown in Figure 1(a). The DRIVER class defines an OONF for an object whose type is more detailed than the PERSON type described in Example 2.3. (I.e., DRIVER contains more attributes than PERSON.) As we discuss in Section 4, the DRIVER class can easily be defined as a subclass of PERSON, allowing most of the DRIVER attributes to be inherited rather than redefined. In addition to the class definitions, the OOBN specifies a situation object and an associated OONF describing the situation of interest. The OONF for *Situation*, whose structure is shown in Figure 1(b), contains the following complex attributes: *Driver*, *Car*, *Weather*, and *Road*, from the appropriate classes. As the figure shows, there are also four attributes which are simple objects. The OOBN must supply CPTs for these objects. ∎

Once we define an OOBN, we can use it to answer arbitrary queries about the objects in our domain. As usual, a query assigns values (evidence) to some objects and inquires about others. Objects which are not at the high level can be accessed using notation such as $X.A$, as described above. For example, one possible query for the accident OOBN is:

*P(Damage = none* |

---
[1] Under the obvious assumption that types in our type system must be finitely nested tuples.

*Driver.Age* = "<20yr", *Road.Location* = *rural*)
Note that this query mentions the object *Road.Location* which is encapsulated in the *Road* object and is therefore not visible within the model. Clearly, even objects that are not visible within the model may be of interest to the user and should be accessible to user queries.

## 3  Inference in OOBNs

Now that we have presented the basic language for defining OOBN models, we turn to the subject of performing inference in OOBNs. The proof of Theorem 2.13 shows how an OONF defines a structure much like a standard Bayesian network. In principle, one could define a Bayesian network over the variables $\vartheta^+(X)$ for every object $X$ in an OOBN. One could then apply a standard BN inference algorithm such as junction trees [Lauritzen and Spiegelhalter, 1988]. Unfortunately, the BN produced in this manner is not structured in a way that supports effective inference. The random variables corresponding to complex objects usually range over a very large set of values, rendering most algorithms impractical.

The key to effective inference in OOBNs is the observation that the complex objects are only used to group together simple objects into coherent units. These higher-level units encapsulate the simple objects, making it easier to define the inputs and outputs of a model in terms of cognitively meaningful entities. However, only simple objects can have a meaningful effect on the model, i.e., by influencing the outcome of some random choice (the value of a random variable). This observation allows us to reformulate OOBN inference in terms of basic variables. Note that this property is specific to the language we have chosen. In a richer language, such as that of [Koller *et al.*, 1997b], the structured objects play a much more important role, preventing us from simplifying the inference algorithm by restricting to simple objects.

For the remainder of this section, let $\mathcal{B}$ be an OOBN, and let $\mathcal{X}$ be the set of objects defined in $\mathcal{B}$. Let $\mathcal{S}$ be the set of simple objects defined in $\mathcal{B}$. Given an assignment $v$ of values to $\vartheta(\mathcal{S})$, we can uniquely reconstruct a value for $\vartheta^+(X)$ for every object $X$ in the model: complex objects are composed out of simpler objects, and for a simple object $S$, $\vartheta^+(S) = \vartheta(S)$. Therefore any distribution over $Val(\vartheta(\mathcal{S}))$ can be extended to a distribution over $Val(\vartheta^+(\mathcal{X}))$. The following lemma allows us to construct a BN over $\vartheta(\mathcal{S})$.

**Lemma 3.1:** *Let $X \in \mathcal{X}$ be an object and $\rho$ an attribute chain in $\vartheta^+(X)$ such that the type of $\vartheta^+(X).\rho$ is basic. Then $X.\rho$ is the is the output attribute of some unique simple object $S \in \mathcal{S}$.*

**Proof:** The proof is by induction on the length of $\rho$. If $\rho$ has length 1, then $X$ must be a simple object, because all value attributes of complex objects are themselves objects, and the type of an object cannot be basic. If $\rho$ is a longer chain $A.\rho'$, then $X.A$ is a value attribute of $X$ and therefore also an object. The inductive hypothesis applied to $A$ and $\rho'$ now implies the claim.  ∎

**Lemma 3.2:** *Let $X \in \mathcal{X}$ be an object, $I$ be an input attribute of $X$, and $\rho$ be a (possibly empty) attribute chain $\rho$. If the type of $X.\rho$ is basic, then $X.\rho$ refers to the output attribute of some unique simple object $S \in \mathcal{S}$.*

**Proof:** The proof is by top-down induction on the structure of objects in $\mathcal{X}$. Since the situation object has no inputs, the base case is trivial. Now, consider some object $X$, an input attribute $I$ of $X$, and an attribute chain $\rho$ such that the type of $X.I.\rho$ is basic. Let Let $Y$ be the enclosing object containing $X$. In the OONF for $Y$, $X.I$ is annotated with some annotation $B.\rho'$. By type compatibility, the output type of $B.\rho'$ must be the same as that of $X.I$. Hence, the type of $B.\rho'.\rho$ is basic. There are now two cases. If $B$ is a value attribute of $Y$, then it is itself an object. By Definition 2.12, $\rho.\rho'$ must be an attribute of $\vartheta(B)$ and therefore also of $\vartheta^+(B)$. Thus, by Lemma 3.1, $B.\rho.\rho'$ is the output attribute of some simple object $S \in \mathcal{S}$. Otherwise, $B$ is an input attribute of $Y$, which is a higher level object than $X$. In this case, the inductive hypothesis applies to $Y$, $B$, and $\rho.\rho'$.  ∎

We can now define a Bayesian network $BN(\mathcal{B})$ which carries the same information as the distribution defined by $\mathcal{B}$. $BN(\mathcal{B})$ contains a node $\vartheta(S)$ for each simple object $S$ defined by $\mathcal{B}$. By the lemma, each input attribute of $S$ must refer to an output attribute of some unique simple object $S'$. We add an edge between $S'$ and $S$ whenever the input attribute of $S$ refers to the output of $S'$. The CPT for $\vartheta(S)$ is given by the OONF for $S$.

**Theorem 3.3:** *$BN(\mathcal{B})$ is a well-defined Bayesian network. The distribution defined by $BN(\mathcal{B})$ over $\vartheta(\mathcal{S})$ is the same as that defined by $\mathcal{B}$.*

**Proof:** We first show that $BN(\mathcal{B})$ is a DAG. We define a lexicographic ordering over objects in $\mathcal{X}$ as follows. The situation object gets the label 1. If the label for an object $X$ is $\sigma$, label the value attributes of $X$ with the labels $\sigma.1, \ldots, \sigma.k$ in a manner consistent with the DAG structure of the OONF for $X$. Suppose $\vartheta(S')$ is a parent of $\vartheta(S)$ in $BN(\mathcal{B})$, and the containing object for $S$ is $X$. Then either $S'$ is an attribute of an object contained in $X$ that precedes $S$, or else $S'$ is an attribute of an input of $X$, in which case it must be an attribute of an object that precedes $X$. Either way, $S'$ must precede $S$ in the ordering.

Since there is an edge from $\vartheta(S')$ to $\vartheta(S)$ precisely when the CPT for $\vartheta(S)$ depends on $\vartheta(S')$, $BN(\mathcal{B})$ is a well-defined Bayesian network. To see that it defines the same distribution over $\vartheta(\mathcal{S})$ as $\mathcal{B}$, consider the nodes of $BN(\mathcal{B})$ in order. Clearly $BN(\mathcal{B})$ and $\mathcal{B}$ define the same distribution over a root node, since they use the same CPT, and it has no parents. For any other node $\vartheta(S)$, the conditional probability in $\mathcal{B}$ of $\vartheta(S)$ given its inputs is defined by the CPT in the OONF for $S$. By the equality constraints on the distribution defined by $\mathcal{B}$, If an input $I$ of $S$ refers to $\vartheta(S')$, $\vartheta(S')$ must be a parent of $S$ in $BN(\mathcal{B})$, and $S.I$ must be equal to $\vartheta(S')$ in the distribution defined by $\mathcal{B}$. Therefore the conditional probability in $BN(\mathcal{B})$ of $\vartheta(S)$ given its parents is the same in both distributions. ∎

**Corollary 3.4**: *For an OOBN $\mathcal{B}$ with situation object Situation, $BN(\mathcal{B})$ induces a probability distribution over $\vartheta^+(Situation)$ which is the same as that defined by $\mathcal{B}$.*

Since $BN(\mathcal{B})$ is a standard Bayesian network, we can use any BN inference algorithms to answer queries. However, we can do even better if we design our inference algorithm to take advantage of the organizational structure encoded in the OOBN. The basic intuition is that most of the attributes of an object are encapsulated within it. Others are passed only to the enclosing object. Only a few attributes have "long-range" influences. Therefore, we would like to design our inference algorithm so that it localizes as much of the computation as possible within objects.

We say that an object $Y$ is *defined in* $X$ if $Y$ is a value attribute of $X$ or a value attribute of an object defined in $X$. A basic variable $\vartheta(S)$ is *used* by an object $X$ if an attribute chain of an input of $X$ refers to the output of $S$. A basic variable $\vartheta(S)$ is *exported* by $X$ if $S$ is defined in $X$ and $\vartheta(S)$ is used by some object not defined in $X$. A basic variable $\vartheta(S)$ is *imported* by $X$ if $S$ is not defined in $X$ but $\vartheta(S)$ is used by some object defined in $X$.

Intuitively, the only information that an object needs to communicate to an enclosing object are the variables that it imports and exports. More formally, we define the *I/O-set* of a complex object to be the set of variables that it imports and exports. As we now show, the I/O-set of a complex object is enough to d-separate the variables corresponding to objects defined in it from the other variables in the model.

**Lemma 3.5:** *Let $U$ be the set of basic variables corresponding to simple objects defined in some complex object $X$. Let $Y$ be $X$'s I/O-set. Then $U$ is conditionally independent of $\mathcal{S} - U$ given $Y$ in $BN(\mathcal{B})$.*

**Proof:** Consider any path between $U$ and $\mathcal{S} - U$. Such a path must contain adjacent nodes $S_1 \in U$ and $S_2 \in \mathcal{S} - U$. There are two possibilities. In the first case, there is an edge from $S_1$ to $S_2$. In such a case $S_1$ is defined in $X$ and used outside of it, so that it is exported by $X$ and therefore in $X$'s I/O-set. Since the path does not have converging arrows at $S_1$, it must be blocked when we condition on $S_1$. In the other case, there is an edge from $S_2$ to $S_1$, so $S_2$ is imported by $X$, and it similarly blocks the path. Therefore the d-separation criterion is satisfied for every path. ∎

This lemma allows us to localize the inference in the graph according to the structure of the objects. More precisely, we can define a set of variables $\Sigma_X$ for each object $X$ in $\mathcal{B}$. The set $\Sigma_X$ consists of the basic variables corresponding to simple value attributes of $X$, the I/O-set of $X$, and the I/O-sets of the complex value attributes of $X$. Thus, $\Sigma_X$ consists of the variables that are local to $X$, as well as any variables "in transit" in either direction (between objects containing $X$ and the objects which $X$ contains). These sets are organized in a tree structure in a natural manner, with the tree-parent of $\Sigma_X$ being $\Sigma_Y$ if $X$ is an attribute of $Y$. We can now construct a separate junction tree for each $\Sigma_X$. Essentially, we make sure that both the junction tree for an object $X$ and the junction tree for its enclosing object $Y$ contain a clique containing all of the variables in $X$'s I/O-set. Since $X$'s I/O set d-separates $X$ from $Y$ (and, in general, from any part of the model not enclosed in $X$), we can simply connect these two cliques in the two junction trees, and get a viable model for the two objects together. This is precisely the process used by Srinivas [Srinivas, 1994] in his work on hierarchical model-based diagnosis.

Unfortunately, even with the locality property of an OOBN, the interfaces corresponding to I/O-sets can still be quite large. We now provide a more efficient construction, based on the *MSBN (multiply-section Bayesian network)* framework of Xiang, Poole, and Beddoes [Xiang et al., 1993]. Their construction follows the same general lines as the simple one described above. However, they show how to construct the various junction trees in a way that allows the clique corresponding to the I/O-set to be decomposed, while still supporting correct probabilistic propagation. As a consequence, their construction results in junction trees with smaller cliques, leading to more efficient probabilistic inference.

Space considerations prevent us from describing the MSBN framework in its entirety. We simply survey some of the basic data structures and their application in our framework. (The simplified definitions are adapted from [Xiang, 1995].) An MSBN partitions the random variables in a BN into a set of non-disjoint *subnets*. Each subnet contains some "localized" set of random variables of the BN, with their associated edges. The intersection between two subnets is called a *d-sepset*, and has the property that it "locally" separates the two subnets. I.e., when the two subnets are considered in isolation of the remainder of the network, their intersection renders them conditionally independent. An *MSBN of hypertree structure* is one in which the subnets are organized into a hypertree. The hyperlinks correspond to the d-sepsets between two adjacent subnets in the hypertree. Each hyperlink has the property that it renders the two parts of the network that it connects conditionally independent. Xiang *et al.* provide an inference algorithm for hypertree MSBNs that guarantees correct probabilistic inference. Given the network $BN(\mathcal{B})$, we define $MSBN(\mathcal{B})$ to contain a subnet $\Sigma_X$ for every object $X$ in $\mathcal{B}$, where $\Sigma_X$ is the set of nodes described above. Consider the tree defined by the set of objects $\mathcal{X}$ in $\mathcal{B}$, where $Y$ is the parent of $X$ if $X$ is a value attribute of $Y$. The subnets of $MSBN(\mathcal{B})$ are organized into a hypertree, in which there is an hyperlink between $\Sigma_Y$ and $\Sigma_X$ when $Y$ is a parent of $X$.

**Theorem 3.6:** *$MSBN(\mathcal{B})$ is an MSBN of hypertree structure.*

**Proof:** Let $Z$ be the set of simple variables contained in the subnets beneath $\Sigma_X$ in the hypertree (including $\Sigma_X$). By the definition of subnet, $Z$ contains all the simple variables defined as well as the IO-sets of objects defined in $X$. Now, let $\vartheta(S)$ be any simple object not defined in $X$, but contained in the IO-set of some object defined in $X$. Then $\vartheta(S)$ must be defined outside $X$, but used by some object defined in $X$, so it is imported by $X$. Hence $\vartheta(S)$ is contained in the IO-set of $X$. Therefore, in the notation of Lemma 3.5, $Z \subseteq U \cup Y$. By the lemma, $Y$ renders $U$ conditionally independent of $S-U$, so it also renders $Z$ conditionally independent of $S - Z$. Hence each hyperlink renders the two parts of the hypertree that it connects conditionally independent, as required. ∎

Figure 2 shows the structure of the MSBN constructed for the accident model. Note that many of the objects appear only within a single subnet, a property induced by the locality structure of the object-oriented model. In a hypertree MSBN, all the message passing in the probabilistic inference process is done along the paths of the tree. For example, in Figure 2, there is no edge between the subnets for *Weather* and *Road* even though they share *Weather.Wetness*. The communication concerning this attribute is passed between these subnets via the containing *Accident Model* subnet.

The efficient localized inference algorithms of [Xiang et al., 1993] apply directly to MSBNs of hypertree structure. They organize the junction-tree according to the OOBN structure, thereby exploiting the separation properties of Lemma 3.5. The decomposition of the junction tree into a tree of smaller junction trees immediately leads to the following result:

**Theorem 3.7:** *For each object $X$ in the OOBN, let $c(X)$ denote the complexity of inference in the junction tree defined by the MSBN algorithm for the subnet of $X$. Then the complexity of the inference in the MSBN defined above is $O(\sum_{X \in \mathcal{X}} c(X))$.*

Thus, the complexity of inference grows linearly with the number of objects in our OOBN. The complexity is exponential in the largest clique used in the junction tree for an individual object. However, if the OOBN is structured in a way that mirrors the locality of the domain, it is unlikely that an object will define very many simple objects, or that its I/O-set will be very large. Thus, we believe that complex junction trees for any individual object will be rare in practice.

Note that the junction tree constructed for an object by the MSBN algorithm may be larger than the optimal junction tree over a stand-alone BN of the same structure, because of the need to have the junction trees of neighboring subnets be compatible with each other. The MSBN construction plays a crucial role in reducing the impact of this problem. As we mentioned above, rather than creating a clique in the junction-tree for the entire I/O-set of an object, MSBNs allow the decomposition of the d-sepset into a sort of junction tree itself. (The cliques in the d-sepset junction tree are called *linkages*.) So long as neighboring junction trees are structured similarly with respect to their shared variables, correct probabilistic inference is possible even without having a single clique as an interface between the junction trees. A full discussion of this issue is beyond the scope of this paper; see [Xiang et al., 1993] for details.

The organizational information provided by the object-structure of the OOBN plays two roles in our construction. First, it helps identify a partition of the BN nodes which is more likely to support locality of inference. It is, of course, possible that a standard BN algorithm would naturally find this partition when creating a junction tree. However, this is not guaranteed to happen, particularly since the task of finding an optimal junction tree is known to be NP-hard. The second role of the OOBN structure is the fact that it allows a straightforward construction of an MSBN. As shown in [Xiang et al., 1993], the d-sepset decomposition provided by an MSBN can result in a data structure which is more efficient than that provided by any single junction tree.

Another advantage of MSBNs arises in repeated interaction with a network. If a user asks repeated queries or incrementally adds evidence about a particular subnet of an MSBN, all computation can be performed locally within that subnet. Evidence asserted on that subnet does not need to be propagated to the rest of the network until the user shifts attention elsewhere. This advantage is very relevant in the context of OOBNs, where a user may often focus attention on one object for an extended period.

Finally, MSBNs are also a useful data structure for reusing computation for multiple objects in the same class. If we have two objects of the same class, their subnets will be essentially identical (up to renaming of variables). We can often use the same junction tree for both objects, avoiding the work of recomputing it.[2] In fact, we can even reuse some of the results of the actual inference process. When an MSBN is first constructed, an initial calibration phase (as in standard junction trees) is used to make the various junction trees internally consistent (calibrating the cliques within a tree) and consistent with each other. Analogously to the junction tree process, this process consists of two phases: collecting beliefs from the entire tree into a single root, and distributing them back. Assume that we conduct this process with the situation object (the root of the object tree) playing the role of the root. Now, consider some object $X$ of class $C$ and the subtree of the MSBN hypertree rooted at $\Sigma_X$. The phase of collecting beliefs within this subtree depends only on the stochastic function of the object $X$ and not on its location within the model as a whole. In particular, the inference process for another object $X'$ of the same class would be identical. By caching (or precompiling) the results of this phase of the process, we can reuse it for other identical objects. Note that this caching process also applies to very complex objects that contain many nested levels of other objects. (E.g., if our model contains two car objects.) In this case, the computation for the entire object, including all the enclosed ones, can be reused. For models involving many complex objects of the same class, the savings can be considerable.

---

[2] It is not always the case that the same junction tree can be used, since the tree's exact structure may depend on the structure of the junction tree of the containing object. However, since objects of the same class are often used in similar contexts, this problem is unlikely to occur very often.

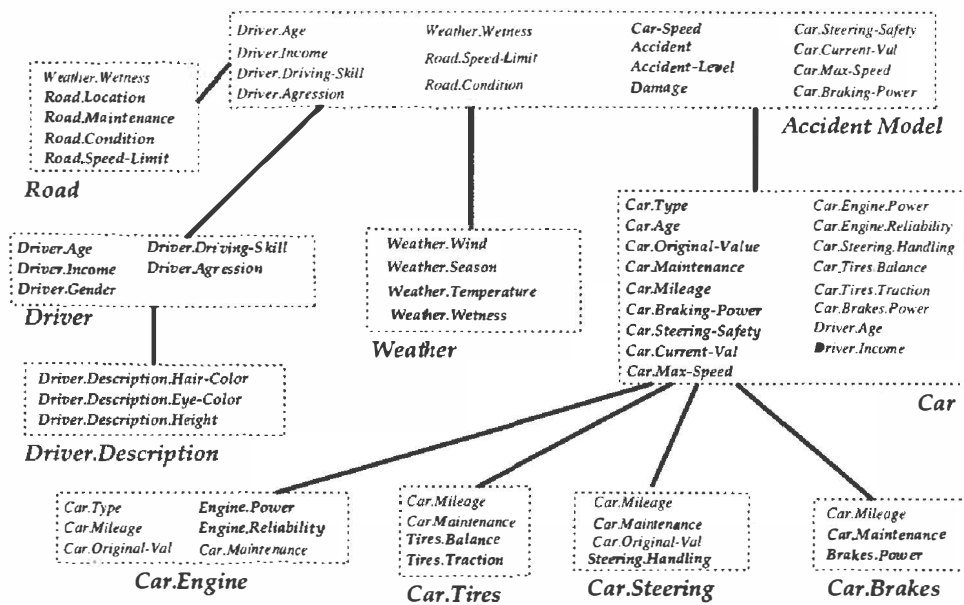

Figure 2: Structure of the MSBN constructed for the accident model. The subnet associated with each object is shown by a box containing the simple objects in the subnet. Objects that are defined locally within a subnet are shown in bold face. There is an edge between any two subnets that interact in the MSBN.

It might seem that the caching can save us at most a factor of 2: In the second phase—distribute beliefs—the beliefs for different identical objects are typically different (based on their location within the model), so we cannot use our results for one object $X$ for another object $X'$. However, this work is not always necessary. Unless we assert evidence about an object or query its internal nodes, there is no need to have its internal beliefs be calibrated. Thus, we can execute a lazy inference process: We leave all objects uncalibrated, and only calibrate them when it becomes necessary. Of course, an inference process such as this requires some bookkeeping in order to remember which calibration operations have not already been done. We defer further discussion to the full paper.

## 4 Subtyping and inheritance

One of the best known and most useful features of object oriented modeling is the ability to create *subclasses* that *inherit* the properties of existing classes. Classes are organized in an *is-a hierarchy*—an instance of a subclass is an instance of its parent class. This inclusion property implies that an instance of a subclass can be used whenever an instance of the parent class is expected. In other words, the subclass must supply all the outputs of the parent class, and cannot require any input that was not supplied to the parent class. Recall that a class is simply an OONF over some set of input and value attributes. The *interface type* of a class $C$ is tuple $\langle I_1 : t_{I_1}, \ldots, I_k : t_{I_k} \to O_1 : t_{O_1}, \ldots, O_m : t_{O_m} \rangle$ where $\mathcal{I}(C) = \{I_1, \ldots, I_k\}$, $\mathcal{O}(C) = \{O_1, \ldots, O_m\}$ and $t_A$ is the type of attribute $A$.

**Definition 4.1:** An interface type $t'$ is a *subtype* of an interface type $t$ if:

- If $t$ has an output attribute named $A$, then $t'$ must have a corresponding output attribute $A$. The output type of $t'.A$ must be a subtype of the output type of $t.A$.

- If $t'$ has an input attribute $A$, then $t$ must have a corresponding input attribute $A$. The type of $t.A$ must be a subtype of the type of $t'.A$. ∎

Let $C$ be a class of interface type $t$ and $C'$ be a candidate subclass of $C$ with an interface type $t'$. The first condition guarantees that, if we replace a $C$ object $X$ with a $C'$ object $X'$, the outputs of $X'$ can be used in place of the outputs of $X$. The second condition guarantees that any inputs that $X'$ requires are available, since $X$ also required them.

At this point, one might define $C'$ to be a subclass of $C$ if the interface type of $C'$ is a subtype of the interface type of $C$. However, this definition is too restrictive. Part of our goal in defining a subclass is to provide additional attributes. While the definition does not prevent that, it does prevent the new attributes from depending on the rest of the model in any new way. Assume, for example, that we are trying to define a COMMUTE-ROAD subclass of the ROAD class. We may want to add a new output attribute *Traffic* to the model. The logical inputs for that attribute are the *Location* attribute, which is internal to the ROAD class, and some attribute representing the time of day. Since there is no appropriate attribute in the ROAD class, we would have to add a new input to the class as a whole, thereby violating the subtyping requirement.

In this example, it seems that the violation of the subtyping property is not a problem. After all, in any situation where we substitute a COMMUTE-ROAD object for a ROAD object, the *Traffic* output is irrelevant: since it does not exist in the ROAD object, it could not have been needed in this context. However, the subtyping assumption, as stated, is required if we want our language to be strongly typed. We finesse our way out of the problem by associating with each class definition a set of different interface types, corresponding to various combinations of input and output attributes, such that the output attributes have all of the required inputs.

Formally, an attribute $A$ in a class $C$ requires another attribute $B$ if $B$ is one of $A$'s ancestors in the DAG of the OONF for $C$. If $\mathcal{O}'$ is a subset of the outputs $\mathcal{O}$ of $C$, the *projected interface type* of C onto $\mathcal{O}'$ is a subtuple of the interface type of $C$, with output attributes limited to $\mathcal{O}'$ and input attributes limited to those input attributes required by $\mathcal{O}'$. The notion of projected type allows us to provide a less restrictive notion of subclass.

**Definition 4.2:** $C'$ is a *subclass* of $C$ if $\mathcal{O}(C) \subseteq \mathcal{O}(C')$, and the projected interface type of $C'$ onto $\mathcal{O}(C)$ is a subtype of the interface type of $C$. An *is-a hierarchy* over classes is a hierarchy $\sqsubset$ over the set of classes such that, if $C' \sqsubset C$, then $C'$ is a subclass of $C$. ∎

The definition of an is-a hierarchy over classes serves two main roles. The first is our ability to use a class as an abstracted (approximate) version of its subclasses. A class and its subclasses may describe the same type of object, but the subclass may be more detailed. For example, we can view the class of PERSON as an abstract class, which has other classes such as STUDENT or DRIVER as subclasses. Each of the subclasses specifies additional details about the object. Furthermore, the distribution specified by a class can be an approximation of the distribution specified by the subclass, but may support more efficient inference. The user can choose the level of detail appropriate to a given problem.

The abstraction property provided by an is-a hierarchy is also useful for simple objects. For example, we may want to refine our INCOMES type, which consisted of {*$0–10K, $10–30K, $30–80K, $80K+*}, into a finer-grained partition: {*$0–10K, $10–30K, $30–80K, $80–130K, $130–250K, $250K+*}. Since we can directly map between these values and the original ones, this new type is a subtype of INCOMES. We can therefore use this new type as our value type for the *Income* attribute of a new class RICH-PERSON ⊏ PERSON.

The second role of the is-a hierarchy is based on the observation that the model for a subclass often has much in common with the model for a superclass. For example, the distribution over the *Description* attribute is the same for the PERSON class and for the RICH-PERSON class. Our class hierarchy can be used to support "code reuse" by allowing subclasses to inherit parts of the specification for the superclass.

More precisely, we allow a subclass $C'$ to be defined by naming a parent class $C$, and then listing any modifications to its definition. By default, all attributes, their types, their OONFs, and their input mapping are inherited from $C$ to $C'$. Additional attributes $A$ in $C'$ are declared as usual, using a type declaration, an OONF, and an annotation for every input attribute in $\mathcal{I}(A)$. The resulting network structure of $C'$ must, of course, continue to be a DAG. If an attribute $A$ is declared in $C$ and an attribute by the same name already exists in $C$, then the $C'$ definition replaces the one in $C$. Of course, the subtype restrictions (Definition 4.1) have to be maintained in this case. (We note that our definition avoids the trap of multiple inheritance by the simple expedient of disallowing it.)

**Example 4.3:** We can augment our class definitions from Example 2.16 with several subclasses that interact with each other in interesting ways. The subclass FUEL-INJECTED-ENGINE ⊏ ENGINE may have a new complex hidden attribute *Fuel-Injection-System*. The *Reliability* attribute is modified to take an additional input, which is mapped to this new attribute. The CPT over FUEL-INJECTED-ENGINE.*Power* gives a higher probability to the value *high*. In the subclass SPORTS-CAR ⊏ CAR, the *Engine* attribute is of type FUEL-INJECTED-ENGINE. The annotations for the inputs of *Engine* are left unspecified, and are therefore inherited from CAR. There is a different CPT over the simple attribute *Original-Value*. There is a new output attribute *Acceleration* with an associated probabilistic model. The attributes *Engine.Power* and *Original-Value* are assigned to the inputs of *Acceleration*. All other aspects of the model are inherited from CAR. ∎

## 5 Refining models

OOBNs provide two orthogonal organizational hierarchies: the *part-of* hierarchy, corresponding to the inclusion of one object within another, and the is-a hierarchy over classes. These two hierarchies combine to provide a powerful and flexible tool for abstraction and refinement. The part-of hierarchy allows us to *iconize* an object, ignoring its component attributes. More precisely, the iconization process defines an object with no hidden variables, but that induces an identical probabilistic model over its interface. The is-a hierarchy allows us to describe an object using a variety of classes, corresponding to different levels of detail in their description. In general, these more abstract classes are also computationally more efficient. Note that iconization naturally fits into the is-a hierarchy: Since the interface type of an iconized object is the same as that of the uniconized version, we can define a class representing the iconized objects, which will then be a (more abstract) superclass of the original class.

These abstraction mechanisms can be applied by the user at runtime to construct models that represent each of the various aspects at the appropriate level of granularity. The user may begin by modeling a domain at a fairly high level of abstraction. Perhaps all the objects will be iconized or represented by high-level classes. As a result of incorporating evidence and querying the model, the user finds certain objects in the model to be particularly relevant. The user can then focus on those objects by deiconizing or refining their models. This process can be repeated, with the user focusing along the part-of hierarchy and refining along the is-a hierarchy.

For example, when modeling an accident domain, it makes sense to begin with iconized versions of car objects. While describing the general circumstances of an accident, the encapsulated attributes of a car are not yet relevant. Inference may be performed more quickly if they are ignored. Suppose the user begins with a model of an accident in which the car is iconized. After asking some queries, that perhaps consider the driver's age and the weather conditions, the user decides to refine the model by specifying that the car is a sports car. After asking some more queries, the user decides to deiconize the sports car, and consider the properties of its brakes.

Consider the work done by the inference algorithm for each refinement. When the car is changed to a sports car, the model remains iconized. The only aspect of the change which is visible to the rest of the model is the probabilistic relationship between inputs and outputs of the car object (the information encoded in the d-sepset between the car object and the rest of the model). This new information needs to be propagated to the rest of the model, but no computation needs to be performed on the internals of the sports car.[3] Later, when the sports car is deiconized, we get the opposite behavior. Evidence about the rest of the model now has to be propagated to the encapsulated attributes of the car. However, the input-output relationship of the sports car object remains the same, so no work has to be done outside of it. This process of propagating information only to certain parts of the model is easily accomodated by the MSBN algorithm.

Suppose instead that the user had deiconized the car before changing it to a sports car. In that case, we already have a representation of the internal probabilistic model of the object. When the model is changed, one might think that the representation needs to be recomputed entirely from scratch. In fact, this is not necessary. The sports car inherits much of its model from the car model. In particular, the models for many subparts of the car are unchanged. Each of these objects is a separate subnet, with its own junction tree. Previous computation done on these objects typically does not need to be changed when the car model is changed.

## 6   Discussion and future work

This paper describes a flexible modeling language for Bayesian networks, based on the object-oriented approach. Of course, OOBNs are not the first proposal for extending Bayesian networks beyond the attribute-based level. However, OOBNs differ from prior proposals in a crucial way. Virtually all of the prior work on this topic focuses on combining Bayesian networks with logic-programming-like rules (see, for example, [Breese, 1992; Ngo et al., 1995; Poole, 1993]). Our approach is based on a stochastic *functional* language [Koller et al., 1997b], with the object-oriented framework a natural extension. We believe that our approach has several important advantages: The ability to naturally represent objects that are composed of lower level objects. And, the ability to explicitly represent classes of objects, crucial for the incorporation of inheritance into the language. These properties are crucial for large-scale knowledge representation. In particular, these mechanisms allow us to reuse model fragments in a way that is natural and semantically coherent, thereby easing knowledge acquisition for complex structured domains. We have also shown that the encapsulation of objects within other objects and the code reuse can provide significant advantages in inference.

Independently of our work, Laskey and Mahoney [Laskey and Mahoney, 1997] have developed a framework for representing probabilistic knowledge that shares some features with OOBNs. In their framework, based on network fragments, complex fragments are built out of simpler ones, in much the same way as complex OONFs are built out of simpler ones. Their framework currently supports the representation of certain features such as hypotheses that we are in the process of incorporating into OOBNs. However, their approach to building complex models is procedural in nature, whereas ours uses a declarative object-oriented representation language. As a result, our approach allows the organizational structure of a model, in particular the encapsulation of objects and the reuse of OONFs within a model, to be expressed explicitly and utilized by the inference algorithm.

These organizational structures also enable one of the most attractive features of OOBNs: its ability to support a natural framework for abstraction and refinement. Currently, the decision as to what level of abstraction to use when modeling a situation is completely up to the user. We are working on mechanisms for automated abstraction and refinement, in which the reasoning system automatically decides the appropriate level of detail to use in answering a query, and incrementally refines the model during the reasoning process.

We are also working on extending the expressive power of OOBNs to allow natural modeling of more complex domains. While OOBNs allow us to utilize the same class hierarchy to define models of a variety of different structures, once a model is described in the language, its structure is fixed. In particular, the language does not allow us to express uncertainty about the identity and number of objects in the model and about the relationships between them. Suppose, for example, that we wanted to consider passengers in the accident model. Since the number of passengers in a car is variable, we would need a different model for each possible number of passengers. A related restriction is that we cannot express global constraints on a set of objects. For example, we cannot say that a car contains three passengers, at least one of whom is a child. The solution is to allow a set of objects to be a type of object. We can then express uncertainty about the number of objects in the set, as well as global properties of the set. Such a language would allow us, for example, to describe a distribution over the number of passengers in a vehicle: a sports car usually has 0 or 1 passengers, while a minivan is likely to have several. Other objects will depend on properties of the set as a whole, rather than on individual objects within the set. In [Koller et al., 1997a], we developed a language and inference algorithm that deals with such sets; we believe the techniques we used can be applied to OOBNs. Of course, there are many other forms of structural uncertainty; we are working on an extension of OOBNs that would allow us to deal with this issue.

---

[3] We note that, while MSBNs do not currently allow part of the model to be changed at runtime, they can be adapted to support this type of interaction.

The other main limitation of OOBNs is the fact that it does not have the expressive power to deal suitably with situations that evolve over time. Objects in an OOBN are static: once an object is defined, its properties are determined once and for all (although we may still be uncertain about them). We would like to be able to apply OOBNs to domains involving multiple interacting entities (e.g., cars) whose state changes over time. One natural model for such a system would define an object for each such entity, with attributes corresponding to its state at different time points. Unfortunately, this type of architecture is not compatible with our current framework. The resulting model is not acyclic, since the different entities influence each other over time. As an alternative, we could define a high level object corresponding to the global state of the system at any given time. While this solution does yield a coherent model, it is inelegant and inefficient, since it requires that we group together the full states of the different objects, thereby breaking their encapsulation. We are currently working on more natural models for modeling dynamic objects.

Despite these limitations, we believe that OOBNs are a significant advance in scaling up Bayesian networks to complex knowledge representation tasks. The key feature of OOBNs that we believe will allow them to scale up is that the representation and the inference go hand in hand. The representation language allows a knowledge engineer to organize a model in a natural and coherent manner. The organization chosen by the engineer contains much knowledge about how the problem decomposes. By following the same organization, the inference algorithm can utilize this knowledge. In essence, where Bayesian networks contain two types of knowledge—relevance relationships and conditional probabilities—OOBNs contain a third type of knowledge—organizational structure.

### Acknowledgements

We would like to thank Kathy Laskey for her great help in making some of our definitions more coherent. We also thank Kathleen Fisher, Uri Lerner, Suzanne Mahoney, David McAllester, Yang Xiang and the anonymous referees for useful comments and discussions relating to this work. This work was supported through the generosity of the Powell foundation, by ONR grant N00014-96-1-0718, and by DARPA contract DACA76-93-C-0025 under subcontract to Information Extraction and Transport, Inc. Some of the work was done while both authors were visiting AT&T Research.